# Visual Anomaly Detection for Reliable Robotic Implantation of Flexible Microelectrode Array

Yitong Chen, Xinyao Xu, Ping Zhu, Xinyong Han, Fangbo Qin*, Shan Yu

*Abstract*—Flexible microelectrode (FME) implantation into brain cortex is challenging due to the deformable fiber-like structure of FME probe and the interaction with critical bio-tissue. To ensure the reliability and safety, the implantation process should be monitored carefully. This paper develops an image-based anomaly detection framework based on the microscopic cameras of the robotic FME implantation system. The unified framework is utilized at four checkpoints to check the micro-needle, FME probe, hooking result, and implantation point, respectively. Exploiting the existing object localization results, the aligned regions of interest (ROIs) are extracted from raw image and input to a pretrained vision transformer (ViT). Considering the task specifications, we propose a progressive granularity patch feature sampling method to address the sensitivity-tolerance trade-off issue at different locations. Moreover, we select a part of feature channels with higher signal-to-noise ratios from the raw general ViT features, to provide better descriptors for each specific scene. The effectiveness of the proposed methods is validated with the image datasets collected from our implantation system.

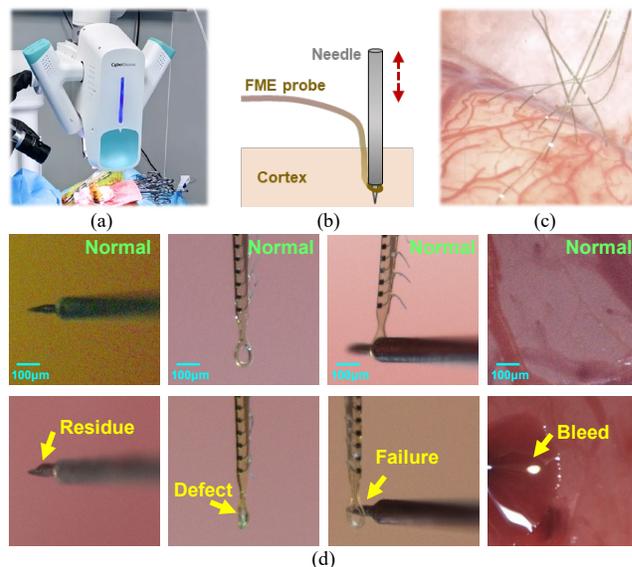

Fig. 1. Robotic FME array implantation. (a) Robotic system. (b) "Sewing machine" paradigm. (c) Implantation results on live macaque cortex. (d) Examples of normal and anomouls cases during implantation.

## I. INTRODUCTION

FLEXIBLE microelectrodes (FMEs) have emerged as key microdevices for implantable brain computer interface (BCI) [1]. The biocompatible polymer material and micrometer-level thickness of FME probe enable safe, precise and long-term neural signal recording after implantation into brain cortex. To achieve high-performance BCI, an array of FME probes is implanted so that the signal channels can be scaled up. A soft FME probe cannot pierce into the cortex by its own stiffness. Hanson *et al*. demonstrated a "sewing machine" implantation paradigm, which utilized a rigid micro-needle to insert soft FME probes into cortex without damaging vasculatures [2]. Following this paradigm, a series of robotic systems have been developed to realize micromanipulation and implantation [3-7]. To fulfill the "sewing machine" like implantation, an FME probe is designed with a loop structure at its end. Under the guidance of microscopes, a micro-needle is precisely controlled to hook the loop structure by threading its tiny tip through the loop. Then the micro-needle pierced into the cortex via a pre-planned position together with the hooked FME probe. Finally, the micro-needle is withdrawn from the cortex alone.

As shown in Fig. 1, we have developed an automated FME implantation robot system, which has successfully realized FME array implantation with live experimental animals. Compared to the basic single FME probe implantation, the continuous implantation of multiple FME probes is more challenging and the reliability is highly emphasized. If anomaly arises but the implantation continues, failure and damage may occur. Therefore, online anomaly detection is essential to ensure the reliability of FME array implantation.

Vision data of a robotic system can be used for execution monitoring and anomaly detection. Inceoglu *et al*. utilized vision and audio data to detect failures in tabletop manipulation tasks with a sensor fusion based network [8]. Ji *et al*. proposed a proactive anomaly detection network for robot navigation in unstructured and uncertain environments [9]. These two methods required both success and failure samples in their training datasets. However, anomalies usually occur with a low probability, making it impossible to enumerate all types of potential anomalies in a limited training dataset.

A more feasible way is to learn the limited normality only from normal cases. Any deviation from normality is detected as an anomaly, which should be addressed immediately to prevent possible risks. Thoduka *et al*. used optical flow and

This work was supported by the National Key Research and Development Program of China (2021ZD0200402) and the National Natural Science Foundation of China (62103413). {*Corresponding author: F. Qin; Email: fangbo.qin@ia.ac.cn*}

Y. Chen, P. Zhu, X. Han, F. Qin, and S. Yu are with the Institute of Automation, Chinese Academy of Sciences, Beijing 100190, China, also with the School of Artificial Intelligence, University of Chinese Academy of Sciences, Beijing 100049, China, also with the State Key Laboratory of Brain Cognition and Brain-Inspired Intelligence Technology, Shanghai 200031, China. X. Xu is with the Sixty-third Research Institution of the National University of Defense Technology, Nanjing Jiangsu 210007, China.

kinematics to predict the motions that occur during a nominal execution. The error between predicted and observed motions was calculated as anomaly score [10]. Sliwowski proposed a vision-language model to learn the preconditions and effects of actions. Anomalies were detected by comparing the expected states defined by a behavior tree to those predicted by the model [11]. However, the methods in [10] and [11] only work for motion- and action-related anomalies, respectively.

The image reconstruction-based anomaly detection strategy trains an autoencoder to reconstruct the normal input. Given an abnormal input, the autoencoder is assumed to result in a large reconstruction residual error. Samuel *et al.* trained a deep autoencoder to detect anomalies like bleeding and blurring in endoscopic surgery vision [12]. Ma *et al.* used an autoencoder to detect camera faults like blur and noise during manipulation tasks [13]. Bozcan *et al.* used a variational autoencoder to detect the layout anomaly of indoor scenes based on grid representation [14]. Zhang *et al.* designed a cascaded reconstruction-discriminant network for accurate localization of industrial visual anomalies [15]. However, the autoencoder training in [12-15] required as many as thousands of normal samples.

The feature embedding-based anomaly detection strategy uses a network to project the image patches into an embedding space, where the distribution of normal and anomalous samples can be compared. This approach is more data efficient because the network can be pretrained on a general large dataset to provide off-the-shelf features. The PaDiM method obtained the multivariate Gaussian distributions of each image patch and used the Mahalanobis distance to measure the anomaly score at each patch location [16]. The patch-specific modeling is highly dependent on image alignment. Differently, the PatchCore method treated the patches at different locations equally and built a memory bank of the features of all the normal patches, then used the nearest neighborhood distance to the memory bank to indicate the anomaly score [17]. PatchCore is robust to image misalignment and performed well on several benchmarks [18,19]. Huang *et al.* proposed a feature registration network that could align the feature maps of various object categories, then compared the registered features to identify anomaly [20].

With our implantation system, observable anomalous cues can be captured by the microscopic cameras. We adopt the feature embedding based strategy considering the data scale. The contributions of our work include: 1) A unified visual anomaly detection framework, MicroVAD, is designed and functions at four checkpoints in each implantation cycle. MicroVAD is also potentially applicable to other microscopic vision guided tasks like cell injection, microassembly, *etc*. 2) Thanks to the system's built-in object localization ability, the regions of interest (ROIs) with aligned anchors are extracted as inputs, so that the misalignment problem is avoided. A progressive granularity patch feature sampling method is proposed to balance the sensitivity-tolerance trade-off at the near-anchor and other locations. 3) Considering the pretrained network provides transferable but non-optimal features for a microscopic scene, we propose a scene-specific feature channel selection method based on channel-wise signal-to-noise ratios (SNR) ranking on normal data.

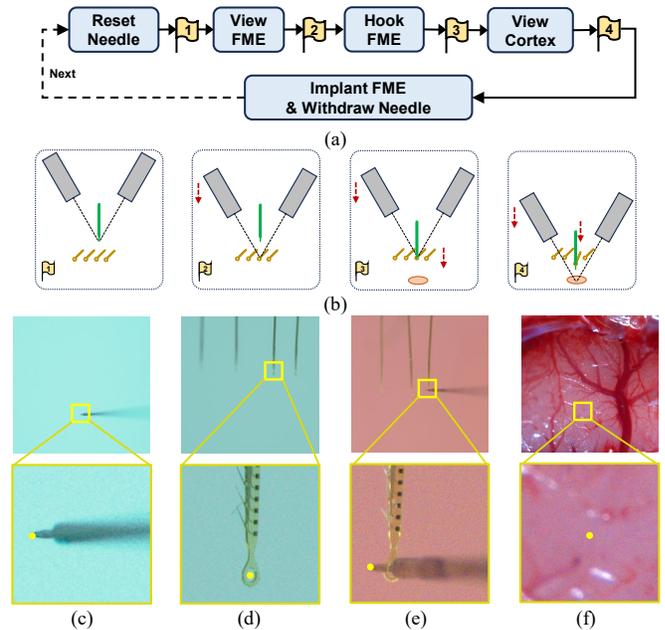

Fig. 2. Four checkpoints (yellow flags) in an implantation cycle (a). The relationships between the microscopes (gray), optical axes (dashed line), micro-needle (green), FME probes (brown), and cortex surface (pink) at the four checkpoints are depicted in (b). The original microscopic images and the anchor-aligned ROIs are shown in (c-f). The yellow dots and boxes indicate the anchors and ROIs, respectively.

## II. METHODS

### A. Task Description

The implantation system has two microscopic cameras, a micro-needle, and a FME array holder. The two microscopes have a narrow common field of view. To focus at different heights, the microscopes can be adjusted by a motorized axis. The micro-needle is driven by high precision motors and has three orthogonal degrees of freedom. The holder is used to temporally fix the FME probes. After the micro-needle hook the loop at a probe's end, the probe can be peeled off the holder with a small force from the micro-needle. As shown in Fig. 2(a), a single probe implantation process includes five steps and four checkpoints.

*Checkpoint 1*: After the micro-needle is reset to the initial position, the needle tip is checked. If anomalous, the needle should be cleaned or replaced. The potential anomalies include deformation, residues, foreign matters, *etc*.

*Checkpoint 2*: The microscopes move downwards to focus on a FME probe's end to check its useability. If anomalous, this probe should be skipped or cleaned. The potential anomalies include structural defects, foreign matters, *etc*.

*Checkpoint 3*: The automatic needle-loop hooking control is executed and the hooking result is checked. If anomalous, the needle is reset and restart the hooking control. The potential anomalies include failure and abnormal termination.

*Checkpoint 4*: The microscopes move downwards to focus on the cortex surface. The target implantation point's eligibility is checked before insertion. If anomalous, manual intervention is required. The potential anomalies include bleeding, unexpected vessels, foreign matters, etc.

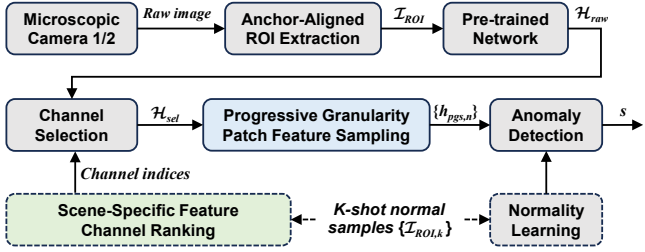

Fig. 3. MicroVAD framework. The modules indicated by dashed line are executed offline.

## B. MicroVAD Framework

The anomaly detection at the four checkpoints are conducted with the unified MicroVAD framework under specific configurations. The framework pipeline is shown in Fig. 3. Since the implantation robot is already able to localize the needle tip and FME loop based on convolutional neural networks (CNNs). After training with adequate labeled images, CNN provides high localization accuracy and strong robustness against disturbance as [21,22] did. The coordinates of implantation point in the cortex surface image is obtained from the planning and navigation system.

We leverage the system's localization results to obtain an anchor-aligned ROI image $\mathcal{I}_{ROI}$ from the raw image. Besides, a predefined offset is added to an anchor to determine the ROI center, so that the ROI contains more foreground. As shown in Fig. 2, the anchors are defined as the needle tip point, loop center point, needle tip point, and planned implantation position at the four checkpoints, respectively.

Since deep networks pretrained on super large dataset can provide off-the-shelf transferable visual features for downstream tasks [23,24], we utilize a pretrained network so that the data-dependent training is avoided. The raw feature map of an anchor-aligned ROI image $\mathcal{I}_{ROI}$ is obtained as $\mathcal{H}_{raw} \in H_0 \times W_0 \times C_0$ tensor. Although the raw feature maps can be directly used for anomaly detection, we plug in the scene-specific feature channel selection and progressive granularity patch feature sampling modules to improve the description ability. The former compresses $\mathcal{H}_{raw}$ as $\mathcal{H}_{sel} \in H_0 \times W_0 \times C_{sel}$ with fewer channels. The latter samples $N_{pgs}$ patch-level descriptors $\{h_{pgs,n}\} \in N_{pgs} \times C_{sel}$ from $\mathcal{H}_{ROI}$ at different location with varying granularities. The transformation from $\mathcal{I}_{ROI}$ to $\{h_{pgs,n}\}$ is denoted as $\{h_{pgs,n}\} = \mathcal{V}(\mathcal{I}_{ROI})$. Finally, the anomaly detection module compares $\{h_{pgs,n}\}$ with the learned normality to estimate the anomaly score $s$. If $s$ is larger than a threshold $\tau$, an anomaly is detected and should be addressed immediately.

Due to the symmetric view angles of the microscopic cameras 1 and 2, their images can share the same MicroVAD configuration at each checkpoint. $K$ normal samples $\{\mathcal{I}_k\}_i$ at the $i^{th}$ checkpoint are collected beforehand, which are used for offline feature mining and normality learning. Considering the cost of various data collection on real platform, only a few normal samples are collected.

## C. Progressive Granularity Patch Feature Sampling

In our preliminary study, we observed that when a model is sensitive to fine-grained anomalies, such as small blood residues on needle tip, it tends to mistakenly identify some normal variations as anomalies. Conversely, when a model tolerates some normal variations, it fails to detect truth subtle anomalies. To address this dilemma, we propose the progressive granularity patch feature sampling method, leveraging the prior knowledge of micro-objects. Near the anchors, i.e. needle tip, probe loop, target insertion point, the features are sampled with a fine granularity, so that the subtle anomaly can be distinguished. In the region far from anchors, such as the background, the features are sampled with a coarser granularity, to lower the sensitivity to normal variations. For this purpose, the granularity factor $g$ at the location $(x, y)$ in a feature map is defined as,

$$g(x,y) = 2^{\min(a_0 + \lfloor d_{anc}(x,y)/\lambda \rfloor, a_1)}, \quad (1)$$

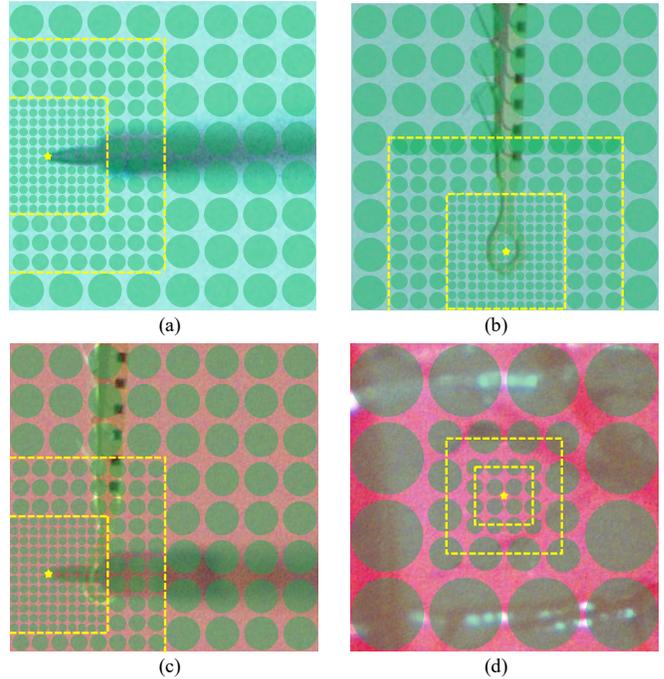

(a) (b)
(c) (d)

Fig. 4. Progressive granularity patch feature sampling. The anchors and distance boundaries are marked by yellow stars and dashed lines, respectively. The granularities at different sampling locations are indicated by green circles. The number of circles equals $N_{pgs}$. Note that the cicle sizes qualitatively indicate the granularity and sampling intervals, not the patch sizes. The equivelant patch sizes of ViT embeddings are larger than sampling intervals.

where $d_{anc}(\cdot)$ indicates the distance from $(x,y)$ to the anchor in $\mathcal{H}_{raw}$. $\lfloor \cdot \rfloor$ is the floor function. $\lambda$ controls the spatial intervals of the granularity changes. $a_0$ and $a_1$ determine the minimum and maximum granularity, respectively. Thus, the granularity progressively becomes coarser as the distance from the anchor to the patch location increases, as shown in Fig. 4.

To suppress the detailed features when the granularity becomes coarser, the spatially average pooling is conducted with $g$ as the kernel size and stride. Considering the low resolution of $\mathcal{H}_{ROI}$, we use the Chebyshev distance instead of the Euclidean distance, namely,

$$d_{anc}(x,y) = \max(|x - x_{anc}|, |y - y_{anc}|), \quad (2)$$

where $(x_{anc}, y_{anc})$ is the coordinates of the object anchor represented in $\mathcal{H}_{raw}$. Spatial average pooling on $\mathcal{H}_{raw}$ with the granularities $\{g_p\}$ as the kernel sizes and strides are used to obtained the pooled feature maps $\{\mathcal{H}_p\}$. Then for each pooled feature map $\mathcal{H}_p$, we sample the features of the patches whose center satisfy (1) after aligned to the original resolution of $\mathcal{H}_{raw}$. Thus, an input image $\mathcal{I}_{ROI}$ generates $N_{pgs}$ patch feature vectors $\{h_{pgs,n}\}$ in total.

### D. Normality Learning and Anomaly Detection

For the $i^{th}$ checkpoint, we collect $K$ normal ROI images $\{\mathcal{I}_k\}_i$ and transform them to $K$ sets of patch features $\{h'_{pgs,n}\}_k$. The normality of the $n^{th}$ patch feature is modeled as a Gaussian multivariate distribution $\mathcal{N}(\mu_n, \Sigma_n)$, where the mean $\mu_n$ and the covariance $\Sigma_n$ can be easily calculated with $K$ normal samples, as given by,

$$\mu_n = \frac{1}{K}\sum_{k=1}^{K} h'_{pgs,n,k}, \quad (3)$$

$$\Sigma_n = \frac{1}{K-1}\sum_{k=1}^{K}(h'_{pgs,n,k} - \mu_n)(h'_{pgs,n,k} - \mu_n)^T + \varepsilon I, \quad (4)$$

where $\varepsilon I$ is a small regularization term to prevent singularity.

Given a test input $\mathcal{I}_{ROIq}$, the patch feature at the $n^{th}$ sampling location is compared with the learned distribution $\mathcal{N}(\mu_n, \Sigma_n)$ based on the Mahalanobis distance, namely,

$$s_n = \sqrt{\left(\mathcal{V}(\mathcal{I}_{ROIq})_n - \mu_n\right)^T \Sigma_n^{-1} \left(\mathcal{V}(\mathcal{I}_{ROIq})_n - \mu_n\right)}. \quad (5)$$

The image level anomaly score $s$ is the maximum Mahalanobis distance among all the $N_{pgs}$ locations, namely,

$$s = \max_n s_n. \quad (6)$$

### E. Scene-Specific Feature Channel Selection

The microscopic scenes in the implantation task are highly structured and involve few object types, while the visual features of the pretrained network are learned from vast amounts of general images. Therefore, the raw features probably have a part of channels that introduce unexpected noises to the distance measure. From only a few normal samples, it is impossible to find a optimal set of feature channels that maximize the testing performance, because the potential anomaly types are assumed arbitrary. A feature channel insensitive to normal objects might be sensitive to an unexpected anomaly element. Conversely, a feature channel that discriminative to normal scenes might be indiscriminative to anomalies. Nevertheless, we can still evaluate the different feature channels with only the $K$ normal samples and select the top $C_{sel}$ channels, to further improve the testing performance.

Inspired by the concept of SNR, we assume that the feature values of the $n^{th}$ patch in $\mathcal{H}_{raw}$ are preferred to have higher response and lower variance under the normal cases. The SNR of the $c^{th}$ channel at the $n^{th}$ patch location over $K$ normal samples is defined as,

$$\beta_{n,c} = \frac{\mu_{raw,n,c}^2}{\sigma_{raw,n,c}^2}, \quad (7)$$

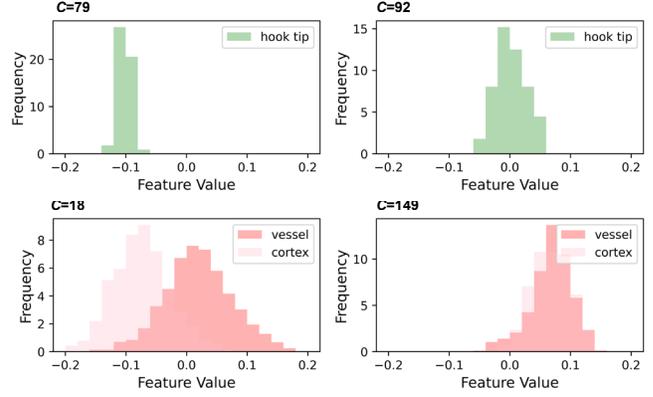

Fig. 5. Distributions of the $c^{th}$ channel of the normalized DINOv2-ViT-S features over the nomal image patches.

where $\mu_{raw,n,c}$ and $\sigma_{raw,n,c}$ are the mean and standard deviation over the $K$ normal samples $\{\mathcal{H}_{raw,n,c,k}\}$, $k=1,2,\ldots,K$. The average SNR $\beta_c$ over all the patch locations is utilized to evaluate the $c^{th}$ feature channel. As an example, in Fig. 5 we visualize the distributions of two feature channels at the locations near the needle tip when the hooking result is normal. Compared to the $79^{th}$ channel, the $92^{nd}$ channel has the lower SNR due to its distribution near zero.

Specially, for the cortex scene at *Checkpoint* 4, the anomalies are mostly related to blood and vessels. The errors in navigation and the uncertainties from bio-tissue might make the insertion point overlapped with prominent vessel and bleeding region. We annotated the vessel and non-vessel patches in the $K$ normal cortex images. Here the feature values are expected to have higher vessel-cortex distinctions and lower intra-category variances. The SNR variant is given by,

$$\beta_c = \frac{(\mu_{cortex,c} - \mu_{vessel,c})^2}{\sigma_{cortex,c}\sigma_{vessel,c}}, \quad (8)$$

where $\mu_{cortex,c}$ and $\sigma_{cortex,c}$ are the mean and standard deviation of the $c^{th}$ feature channel over all the annotated normal cortex patches. $\mu_{vessel,c}$ and $\sigma_{vessel,c}$ are the mean and standard deviation of the $c^{th}$ feature channel over all the annotated normal vessel patches. As an example, in Fig. 5 we visualize the distributions of two feature channels at the vessel and non-vessel locations. Compared to $18^{th}$ channel, the $149^{th}$ channel has the lower SNR because the valid distinctions are not provided.

Finally, the channel-wise SNR scores are ranked and the top $C_{sel}$ channels are picked out for normality learning and anomaly detection. In this work, we set $C_{sel}$ as 50% of $C_0$.

## III. EXPERIMENTS

### A. Hardware Configurations and Evaluation Datasets

The microscopic cameras had the 0.6mm optical depth, the 2048×2448 pixels resolution, and the 2.3μm equivalent pixel size. The ROI size was 256×256 pixels. The tungsten micro-needle had a T-shaped end with the diameter of ~30μm. The FME probe was about 100μm in width and 8μm in thickness. The diameter of the loop structure was ~50μm. The datasets of the four checkpoints were collected from the past implantation processes with live animals and simulated tissues, which involved different micro-needles and FME probes of the same type. The animal experiments were conducted under the ethical review of the Institute of Automation, Chinese Academy of Sciences. The MicroVAD framework ran on the upper computer with a Nvidia RTX 3090 GPU and the ROS noetic system.

TABLE I
ANOMALY DETECTION PERFORMANCES WITH METRIC AUPR (%)

| | Methods | Needle | FME | Hook | Cortex |
|---|---|---|---|---|---|
| | PaDiM [16] | 96.20 | **96.12** | 82.32 | 61.80 |
| | PatchCore [17] | 96.33 | 80.64 | 68.32 | 66.34 |
| | CAReg [20] | 94.31 | 76.92 | 88.54 | 76.76 |
| MicroVAD | CNN+without PGS/FCS | 93.30 | 92.97 | 84.52 | 50.43 |
| | ViT+ without PGS/FCS | 97.04 | 91.05 | 91.08 | 69.11 |
| | ViT+PGS+without FCS | 97.48 | 94.11 | 94.83 | 91.43 |
| | ViT+PGS+FCS (bottom-50%) | 95.36 | 90.53 | 92.42 | 88.94 |
| | ViT+PGS+FCS (top-50%) | **98.28** | 95.90 | **95.78** | **93.28** |

TABLE II
PERFORMANCE EVALUATION OF THE PROPOSED MicroVAD
AFTER ANOMALY SCORE THRESHOLDING

| Metrics | Needle | FME | Hook | Cortex | Average |
|---|---|---|---|---|---|
| F1-score-max | 0.948 | 0.898 | 0.893 | 0.870 | 0.902 |
| Recall (%) | 100.0 | 100.0 | 93.59 | 0.870 | 95.15 |
| Precision (%) | 90.20 | 81.48 | 85.38 | 0.870 | 86.01 |

For each checkpoint, $K=8$ normal images were used for normality learning and feature channel selection. The micro-needle and FME probe images had axial symmetry so that we flipped them vertically and horizontally for data augmentation, respectively. The cortex images had central symmetry so that we flipped them in arbitrary direction for data augmentation. As a result, 16, 16, 8, and 32 normal images were used for normality learning and feature channel selection. The test dataset of *Checkpoint* 1 contained 12 normal and 46 anomalous samples. The test dataset of *Checkpoint* 2 contained 32 normal and 44 anomalous samples. The test dataset of *Checkpoint* 3 contained 76 normal and 156 anomalous samples. The test dataset of *Checkpoint* 4 contained 83 normal and 69 anomalous samples.

B. Comparison Experiments

A series of experiments were carried out to investigate the anomaly detection performances with the area under precision-recall curve (AUPR) as the metric. The MicroVAD with different configurations were evaluated. The DINOv2-ViT-S network [24] pretrained on the LVD-142M dataset was used as the feature extractor. The patch size and strides were 14 and 7 pixels, respectively, so that the input image was slightly resized to 252×252 pixels to be compatible with the patch size. Optionally, the ResNet-50 network [25] pretrained on the ImageNet dataset was also tested, with its second block's output as the raw feature map. The parameter triplet $<\lambda,a_0,a_1>$ in (1) was set as $<6,0,2>$ for the first three checkpoints and $<3,1,3>$ for the fourth checkpoint. Besides, the existing fea-ture embedding-based anomaly detection methods PaDiM [16], PatchCore[17], and CAReg[20] were compared based on their open source codes, which used Wide-ResNet50, Wide-ResNet50, and Feature Registration network as the feature extractors, respectively. The subsampling rate in PatchCore was empirically set as 20%. The estimator in CAReg (hazelnut) adopted the Mahalanobis distance. The flipping based data augmentation were also involved in these methods. The AUPR scores are reported in Table I.

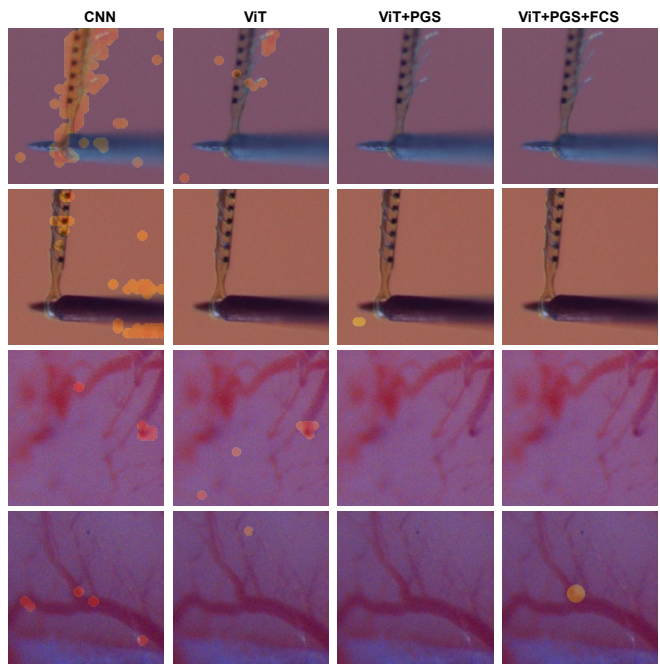

Fig. 6. Comparison of different configurations of MicroVAD. The anomaly JET colormap is overlapped on the input image for visualization. The circle markers indicate the anomaly patches exceeding the thresholds determined by the maximum F1-scores. The first two rows are normal hooking results. The third row is a normal case where the implantation point at the center is vessel-free, while the fourth row is a anomalous case where the implantation point is near a prominent vessel.

*1) ViT vs CNN*: We tested the vanilla version of MicroVAD without using progressive granularity patch feature sampling (PGS) and scene-specific feature channel selection (FCS). Comparing to ResNet-50, DINOv2-ViT-S provided the significantly superior performances on the Needle, Hook, and Cortex images. The performance gap on FME images was relatively small. As shown in Fig. 6, the CNN features led to more false positives because they mainly encoded textures and local shapes, so that the invariance against the deformations of flexible probes and vessels was relatively low, especially when only a few normal samples were given.

*2) Effectiveness of PGS*: The PGS module contributed the AUPR improvements of 0.44%, 3.06%, 3.75% and 22.32% on the four test sets, respectively. The key benefits of PGS stemmed from the different sampling granularities of the patches near and far from the anchor. As shown in Fig. 6, using a uniform sampling granularity would cause more false positives in the less important regions far from the needle tip and center implantation point.

*3) Effectiveness of FCS*: After integrating FCS, the AUPR scores were further improved by 0.80%, 1.98%, 0.95%, and 1.85% on the four test sets. To prove that the effectiveness of feature channel selection was not random, we also tested the feature channels ranking the bottom 50%, which resulted in a significant performance deterioration. As shown by the second and fourth rows in Fig. 6, the selected feature channels provided no noisy false positive around the needle tip and better discriminability for vessel near the center.

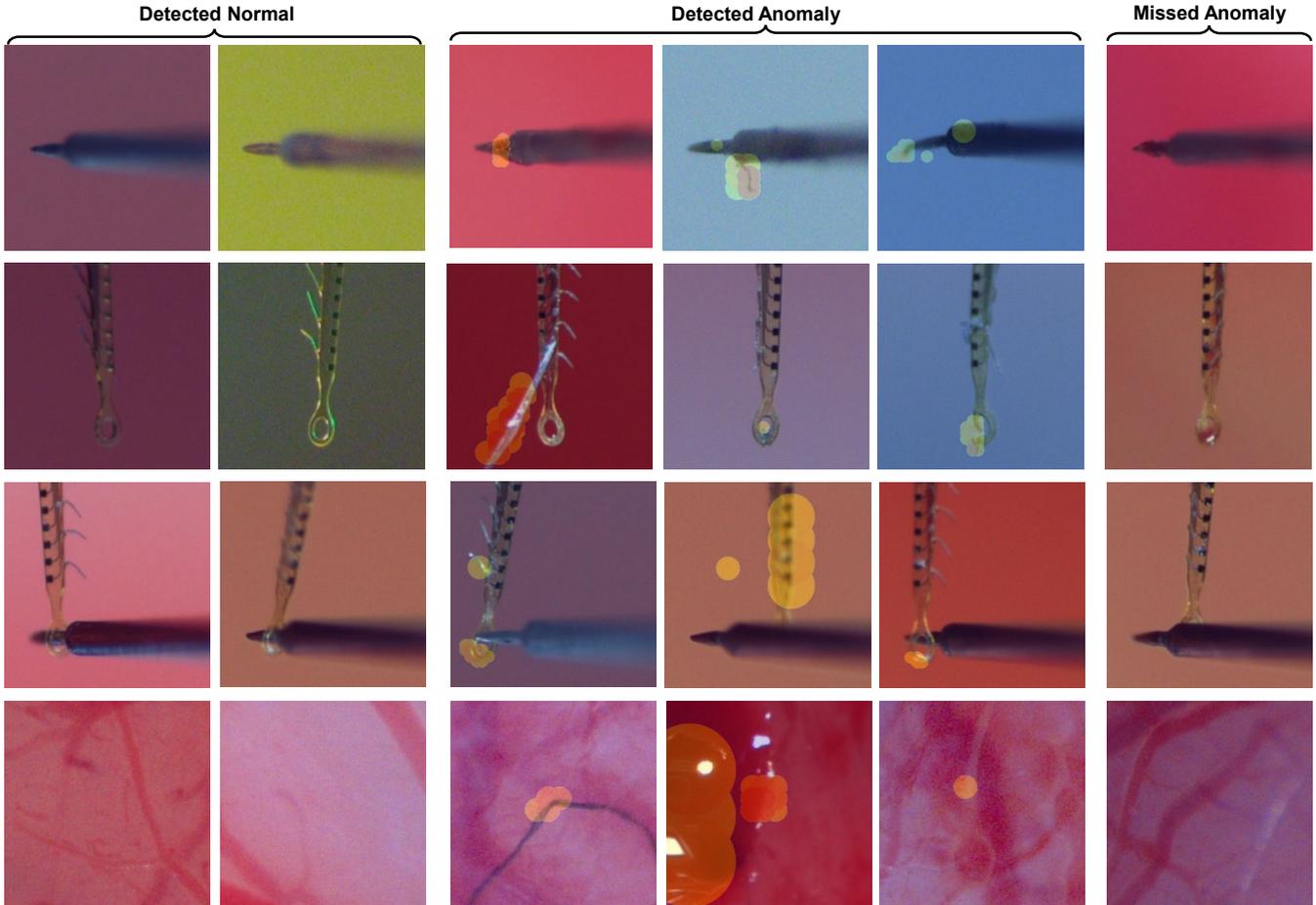

Fig. 7. Visualized anomaly detection results of test samples with the proposed MicroVAD. The anomaly JET colormap is overlapped on the input image. The circle markers indicate the anomaly patches exceeding the thresholds determined by the maximum F1-scores

*4) Comparative Methods:* PaDiM had the best AUPR score on FME images, however its performance on Hook images was not satisfactory. The proposed MicroVAD with PGS and FCS provided the best AUPR scores on Needle, Hook, and Cortex images, while its performance on FME images was comparable to that of PaDiM. Therefore, the comparative methods originally designed for industrial scenes did not directly match our task with microscopic scenes and limited normal samples.

## C. Performances after Thresholding and Limitations

Using the thresholds corresponding to the maximum F1-score, the recalls and precisions were obtained as shown in Table II. The average recalls and precisions on the four test sets were 95.15% and 86.01%, respectively. The visualized anomaly detection results of test samples are shown in Fig. 7. Overall, the proposed MicroVAD was able to detect various anomaly types while tolerating the normal variations. The average execution time of MicroVAD was 16.5ms. Some failure cases, where subtle anomalies occurred, are also shown in Fig. 7, including the little residue on needle tip, the blocked loop structure, the failed hooking result which seemed like normal except for the total occlusion of the tiny loop, and the implantation point near the edge of a prominent vessel.

## IV. CONCLUSION

Towards reliable robotic implantation of multiple FME probes, this paper proposes a unified visual anomaly detection framework for the four critical checkpoints in an implantation cycle. The micro-object ROIs with aligned anchors are extracted from the original microscopic images, with the built-in object localization functions. The progressive granularity patch feature embedding methods was proposed to analyze the patches at different locations with varying sampling granularity. The feature channels of the raw ViT embeddings are ranked according to their SNR scores evaluated on normal samples of a specific scene, and the top-ranking channels are selected to serve as the scene-specific visual features. The proposed framework demonstrated promising performances under the challenges of limited normal data and various anomaly types. The future work will attempt to introduce more task prior knowledge and the semi-supervised learning to improve the sensitivity to subtle anomalies while guaranteeing the tolerance to normal variations.